%% file: main.tex
\documentclass[pdflatex,sn-vancouver-ay]{sn-jnl}
\usepackage{graphicx}%
\usepackage{multirow}%
\usepackage{amsmath,amssymb,amsfonts}%
\usepackage{amsthm}%
\usepackage{booktabs}
\usepackage{array}
\usepackage{mathrsfs}%
\usepackage[title]{appendix}%
\usepackage{textcomp}%
\usepackage{manyfoot}%
\usepackage{booktabs}%
\usepackage{algorithm}%
\usepackage{algorithmicx}%
\usepackage{algpseudocode}%
\usepackage{listings}%
\usepackage{hyperref}
\usepackage{multirow}
\usepackage[table,xcdraw]{xcolor}

\theoremstyle{thmstyleone}%
\theoremstyle{thmstyletwo}%
\theoremstyle{thmstylethree}%

\begin{document}

\title[Taggus]{Taggus: An Automated Pipeline for the Extraction of Characters' Social Networks from Portuguese Fiction Literature}

\author[1]{\fnm{Tiago G.} \sur{Canário}}
\equalcont{These authors contributed equally to this work.}
\author[1]{\fnm{Catarina} \sur{Duarte}}
\equalcont{These authors contributed equally to this work.}
\author*[1]{\fnm{Flávio L.} \sur{Pinheiro}}\email{fpinheiro@novaims.unl.pt}
\author*[2]{\fnm{João L. M.} \sur{Pereira}}\email{joao.pedro.pereira@uevora.pt}

\affil[1]{\orgdiv{NOVA Information Management School (NOVA IMS)}, \orgname{Universidade Nova de Lisboa}, \orgaddress{\street{Campus de Campolide}, \city{Lisboa}, \postcode{1070-312}, \country{Portugal}}}
\affil[2]{\orgdiv{Centro Algoritmi/LASI}, \orgname{University of Évora}, \orgaddress{\city{7000-671}, \postcode{Évora} \country{Portugal}}}


\abstract{Automatically identifying characters and their interactions from fiction books is, arguably, a complex task that requires pipelines that leverage multiple Natural Language Processing (NLP) methods, such as Named Entity Recognition (NER) and Part-of-speech (POS) tagging. However, these methods are not optimized for the task that leads to the construction of Social Networks of Characters. Indeed, the currently available methods tend to underperform, especially in less-represented languages, due to a lack of manually annotated data for training. Here, we propose a pipeline, which we call Taggus, to extract social networks from literary fiction works in Portuguese. Our results show that compared to readily available State-of-the-Art tools --- off-the-shelf NER tools and Large Language Models (ChatGPT) --- the resulting pipeline, which uses POS tagging and a combination of heuristics, achieves satisfying results with an average F1-Score of $94.1\%$ in the task of identifying characters and solving for co-reference and $75.9\%$ in interaction detection. These represent, respectively, an increase of $50.7\%$ and $22.3\%$ on results achieved by the readily available State-of-the-Art tools. Further steps to improve results are outlined, such as solutions for detecting relationships between characters. Limitations on the size and scope of our testing samples are acknowledged. The Taggus pipeline is publicly available to encourage development in this field for the Portuguese language.
}


\maketitle
\section{Introduction}
Digital Humanities are increasingly crucial in the field of literary analysis \citep{trovati2014extraction, ganascia2015logic, porter2023emotion, amalvy2024interconnected, StanfordLiteraryLab,chakraborty2021detection}. In fact, by leveraging advanced computational approaches to collect, organize, model, and analyze large volumes of data from literary works, researchers can revisit previously unexplored theories and propose new angles on old paradigms. For example, \cite{elson2010social} were able to dispel the idea that novels taking place in rural environments would contain fewer and closely related characters than in novels taking place in urban settings, which would have more characters and also be more complex. A quantitative, large-scale analysis does not support this long-standing dogma in the literary field, and \cite{gessey2020narrative} further demonstrates how complex and densely populated fictions can mimic real-life social networks to become cognitively accessible to readers.

Hence, automating the extraction of characters enables researchers to apply methods and theories from Social Network Analysis to deepen our understanding of plot mechanics and their realistic representation of real-world scenarios. Indeed, extracting social networks of characters has become a priority in the field due to its ability to tackle higher-level tasks and conduct analysis in exceptionally large corpora \citep{moretti2011literary}. An example of the type of tasks that can be performed with this automation is the study conducted on the sacred texts of the Chinese Buddhist canon that consists of fifty-five volumes and over two thousand additional texts, and, as such, becomes impractical for any human-conducted analysis \citep{lee2016conversational}. 

However, applying conventional off-the-shelf Natural Language Processing (NLP) techniques to automate the extraction of characters and their relationships in ambiguous and diverse subjects, as present in books of fiction, has proven particularly challenging \citep{rocha2014entity}. Current techniques are neither optimized nor designed to treat such texts. Hence, holistically, they achieve suboptimal results in the necessary tasks for the Character Network Extraction process -- such as character identification and unification, interaction detection, and graph extraction -- when compared to customized techniques \citep{labatut2019extraction}. 

In the particular case of the Portuguese language, past works have used heuristics to find instances of characters speaking in children’s stories \citep{mamede2004character}, performed the extraction of information on characters presented in novels \citep{bick2023extraction}, or performed the general entity extraction on non-fictional books \citep{rocha2014entity}. To the best of the author's knowledge, the only Character Network Extraction System for Portuguese depends on external information sources, namely Wikipedia, to extract the characters present in a novel \citep{silva2023analyzing}. As such, it is limited to the small universe of Portuguese novels on Wikipedia and the corresponding major characters in the plot mentioned in those texts, leaving out other characters. Currently, there are no systems that fully automate the Character Network Extraction process for Portuguese literature.

Here, we propose a pipeline Taggus that was designed to perform the automatic extraction of Character Social Networks from books of fiction written in Portuguese. We show that Taggus outperforms existing tools by $50.7\%$ in the task of character extraction and co-reference resolution and by $22.3\%$ in interaction detection. In that sense, the manuscript is organized as follows, we start with a (i) thorough analysis of existing solutions in English but also in less-represented languages such as French and German; (ii) the description of the Taggus pipeline built by levering previously studied heuristics and other ruled-based approaches to the Portuguese language; (iii) we present a new corpus of manually tagged chapters to assess the pipeline’s performance in Portuguese novels; and (iv) an evaluation process that compares the new pipeline against readily available state-of-the-art tools (SOTA) tools that can perform the different subtasks.

The Taggus pipeline is public and available on Github \citep{Taggus}.

\section{Related Work}\label{sec:literature_review}
In literary studies, Social Network Analysis \citep{borgatti2024analyzing} offers the possibility of using graphs as a means to model/represent the web of relationships between characters -- i.e., characters are represented as nodes/vertices and social relationships as links/edges connecting pairs of nodes \citep{grayson2016sliding} -- and, thus, relate how social structures contribute to drive the plot forward \citep{silva2023analyzing, elsner2012kernels, moretti2011literary, jayannavar2015validating}.

There are three main lines of research in which Social Network Analysis intersects with literary studies \citep{labatut2019extraction}: (i) Narrative Analysis, (ii) Social Network Analysis, and (iii) Automation of Character Networks Extraction. Narrative Analysis focuses on a small number of narratives at a time. It is performed to understand a narrative through its plot better, closely related to how the characters interact \citep{colladanay2015clustering}. The goal is to compare novels from the same author or period \citep{Rieck2016Networks} as well as test standing dogmas of literary studies \citep{elson2010social}. There are also reports of these works being used in educational settings \citep{bolioli2013betrothed}.

Social Network Analysis focus on characterizing the complex nature of relationships between characters and to map their degree of similarity/mimicry of real-world social networks \citep{Alberich2002MarvelUL, bossaert2013harry, everton2022strong, gessey2020narrative,chakraborty2021detection}, upholding the idea that understanding these networks, even if fictitious, can help us better understand real-world networks \citep{maccarron2012universal,trovati2014extraction}. In these two lines of research, researchers commonly study small samples of books and manually curated/annotated information \citep{Rieck2016Networks}.

The third line of research, Automation of the Character Networks Extraction, leverages techniques from text mining and natural language processing to automate the task of entity recognition (i.e., extraction) and consolidation (i.e., co-reference resolution) of social networks from literary works. Hence, the goal is to enable larger-scale projects and analysis, reproducibility of results, and turning an extremely time and resource-consuming process into an efficient and accessible one \citep{moretti2011literary}.  

It is important to note that these lines are not exclusive. One work can fit into two or more lines, as automation projects are usually built to conduct the previously mentioned analysis and test hypotheses \citep{elson2010social}.

\subsection{Automation of the Character Network Extraction}\label{sec:related_work_network_extraction}
Literary novels present unstructured text \citep{lee2016conversational} that is opaque in both form and meaning, proving to be challenging to even human understanding \citep{zhai2013web, labatut2019extraction}. A straightforward solution is to apply Named Entity Recognition (NER) tools, a typical task of NLP, to extract characters' names from plain text. Traditionally, NER research focuses on finding entities in text, such as persons, locations, or organizations. While less traditional, it can also be tuned to other less general entity types like gene names \citep{sarawagi2008}. Character names extraction can be seen as a simplification of NER or even a specialization, as we are only interested in a specific type of person. State-of-the-art NER \citep{yamada-etal-2020-luke,wang-etal-2021-automated} are language model-based solutions that are trained over annotated texts on specific domains.

The performance of NLP methods in performing essential tasks for Character Network Extraction, such as Named Entity Recognition (NER), is often suboptimal \citep{bornet2017character}. The reason is that NLP models are commonly trained with text sources such as news articles and social media posts, contrasting with the self-contained "worlds" portrayed in novels and the unique linguistic styles of each author \citep{dekker2019ner}. To address such challenges, previous authors have relied on large manually annotated corpora of literary novels, developed either as a collective effort from academic studies \citep{vala2015mr,dekker2019ner} or achieved by crowdsourcing non-expert annotations \citep{callison2010mechanical,jha2010corpus,yuen2011survey} to train and evaluate their models. However, these efforts have mainly concentrated around the English language, creating a significant disparity between the state-of-the-art tools in English and in other less-represented languages \citep{dasilvaconrado2014survey}. 

Hence, outside English, the lack of reliable NLP models and the specifications of each language --- sentence structure, name and title rules --- lead to a prevalence of rule-based approaches for the automation of character networks extraction in languages \citep{bornet2017character, Besnier2020HistoryTM}. These procedures benefit from the ability to understand the logic behind the results and produce refinements to the pipeline that can be quickly adjusted in case of need \citep{krug2015rule}.  That is the case for past works on the Portuguese language \citep{dasilvaconrado2014survey}, where previous efforts have focused on part-of-speech (POS) tagging and heuristics catered to the Portuguese names \citep{silva2023analyzing}. 

Past works have explored pipelines specific for the task of Character Network Extraction in Portuguese language. Initially, addressing direct discourse in children’s stories \citep{mamede2004character}, then in general entity extraction from non-fictional books \citep{rocha2014entity}. More recently, the extraction of information on characters present in a novel \citep{bick2023extraction}, such as family relations and professions,  and full pipelines for Character Network Extraction that rely on annotations and databases from external sources such as Wikipedia to retrieve character names and relations \citep{silva2023analyzing}. Naturally, such reliance on external sources limits the generalization of the task, and makes the pipeline highly susceptible to the quality of data presented in such external sources, leading to missing character names.

To the best of the authors' knowledge, there is still a need for an automatic character network extraction system for Portuguese literature that can be applied to a large corpus of novels. 

\section{A Standard Pipeline and Theoretical Challenges}\label{sec:related_work_system}
Pipelines used to extract network information from text sources commonly follow a three-step architecture \citep{labatut2019extraction}
: (i) extraction of entities (in this work, character names) and co-reference resolution; (ii) interaction detection; and (iii) generation of the network visualization. These steps can be used to extract different types of networks, depending on the type of relationships being studied, for instance, conversational networks \citep{he2010actor} or friendship relationships (e.g., friend or foe) \citep{srivastava2016inferring}. 
In this section, we review the common elements used in each of the three steps.

The first step consists of extracting the characters' names present in a document (i.e., the novel). That is, not only all the different characters but also all variations in the name of each character. A process designated by co-reference resolution \citep{vala2015mr,bhattacharya2005relational} addresses that the same character can be mentioned using different names, nicknames, or even titles by grouping all the different names. Additionally, the frequent occurrence of family members that share the same last name or, in some cases, can have more than one name in common adds complexity to the task. The variance of possible mentions that can be used for the same character, combined with the use of pronouns used during the novel and anaphoric mentions such as “the doctor” and “the father”, poses many barriers to the automation of this process. Consequently, past works have often considered only nouns, excluding any nominals and pronouns with the justification that discarding these does not lead to a loss of information \citep{labatut2019extraction}. Manually annotated training corpora can provide NLP models with the means to solve some of the naming ambiguity \citep{aroyo2013crowdtruth,sabou-etal-2014-corpus}

Name Entity Recognition (NER) methods are commonly used to identify entities present in the text, including not only persons but also companies and locations. However, these do have limitations in less-spoken languages. As such, a trend has developed where a predefined list of characters, either curated manually \citep{he2013identification} or obtained from external sources \citep{shahsavari2020automated,silva2023analyzing}, is matched to the names present in the text. It is crucial to note that this is unsuitable for automation since books from less-spoken languages may not have this data type available in some available data source (e.g., Wikipedia), and manually curating name lists does not scale. A more consistent approach that has been employed and shown promising results in languages other than English, utilizes Part-of-speech (POS) tagging \citep{rocha2014entity,okeefe-etal-2012-sequence}. This POS-based methods takes advantage of other attributes other than just nouns, for instance gender and honorifics --- such as “Sir” or “Lady” --- and double checks for abidance between gender of characters and honorifics \citep{colladanay2015clustering}; but also post-processing the resulting list of character names, by redacting names that are outliers or appear less than a pre-determined number of times \citep{elson2010social}.

The second step consists of detecting interactions between characters, or social relationships. What counts as an interaction varies based on the specific problem being addressed, and there is ongoing debate about the most appropriate method for capturing the most accurate Social Network \citep{colladanay2015clustering}.

There are five different approaches to what counts as an interaction between characters in Literary novels.
Co-occurrence, where interaction is assigned when two characters appear in a pre-determined unit of text, e.g., a sentence \citep{colladanay2015clustering}. Conversations, where only verbal interactions are marked as such, this strategy works best for theatre plays since most action is portrayed through speech \citep{elson2010attribution,chak2017identifying}. However, for novels, many interactions between characters can be described in the body of text and thus are lost \citep{min2019modeling}, with the advantage being the possibility for directed networks that co-occurrence does not allow for \citep{moretti2011literary}. In some specific cases, authors aim not for the standard Social or Conversational Networks that models interactions but instead hope to map the affiliations between the characters present in the Novel \citep{valls2021automatic}, being the nature of the relationship, such as enemies or allies \citep{srivastava2016inferring} or family relations \citep{bick2023extraction}. Another hypothesis that has received some attention posits that character interactions can be identified through direct actions \citep{mamede2004character}. In these cases, interactions are attributed to certain verbs that can be found in written text that describe interactions between characters, such as “talked", "fought", "helped”, and two characters are assigned an interaction when one of these verbs is found between them. The last approach involves combining the previously mentioned techniques to maximize their effectiveness, such as identifying character dialogues and conversations that utilize specific verbs, thereby leveraging both conversational and direct action approaches \citep{he2010actor,agarwal2013sinnet}.

In the context of Taggus, a co-occurrence approach is chosen, since it seems to be the most appropriate method to detect interactions; it is, however, relevant to note the theoretical challenges it poses. This catch-all method is, by definition, imprecise, leading to false-positive interactions. However, there is proof that false negatives are not as common, and a case can be made that, if two names co-exist in a sentence, even if it does not express a direct interaction between two characters, it most likely reveals some form of connection between them. Therefore, it is not a hurtful consequence when mapping out social networks of characters \citep{colladanay2015clustering}.

The third and final step involves building and rendering the social network between characters based on the extracted information. Past works have considered that the goal of character networks is to represent the entirety of the social structure of the novel characters, thus allowing for the comparison between different novels/authors. As such, networks are considered static projections that ignore any temporal dynamics associated with the character’s social interactions throughout the story \citep{oelke2012advanced}. Some authors have questioned how well static networks can capture evolutions and changing relationships between characters throughout a story \citep{agarwal2012social} and opt instead for dynamic networks. These operate by dividing the narrative into multiple time frames and capturing the social networks from each of these windows, with the most common time unit being the chapters \citep{colladanay2015clustering}. 
For this work, the networks are created over the complete book; however, it is straightforward to generate a timeline per chapter, as this information is well encoded in the data.

\section{Manually Annotated Dataset}

\input{Table1}

We evaluate the proposed Character Network Extraction pipeline, Taggus, on a manually annotated corpus of eleven Portuguese-language novels sourced from Projecto Adamastor\footnote{https://projectoadamastor.org/}. The corpus comprises six Portuguese and five Brazilian novels by ten different authors, published between 1862 and 1941, as seen in Table~\ref{tab:list_novels}. This diversity test tags the Taggus pipeline across various writing styles and is not skewed by overfitting to a specific author’s writing \citep{elson2010social}.

A pre-processing step was performed in the corpus to remove non-literary text items such as publishing or distribution information and author notes to ensure that only literary text is considered while performing our information extraction process \citep{dekker2019ner}.

\begin{figure}[!t]
    \centering
    \includegraphics[width=\textwidth]{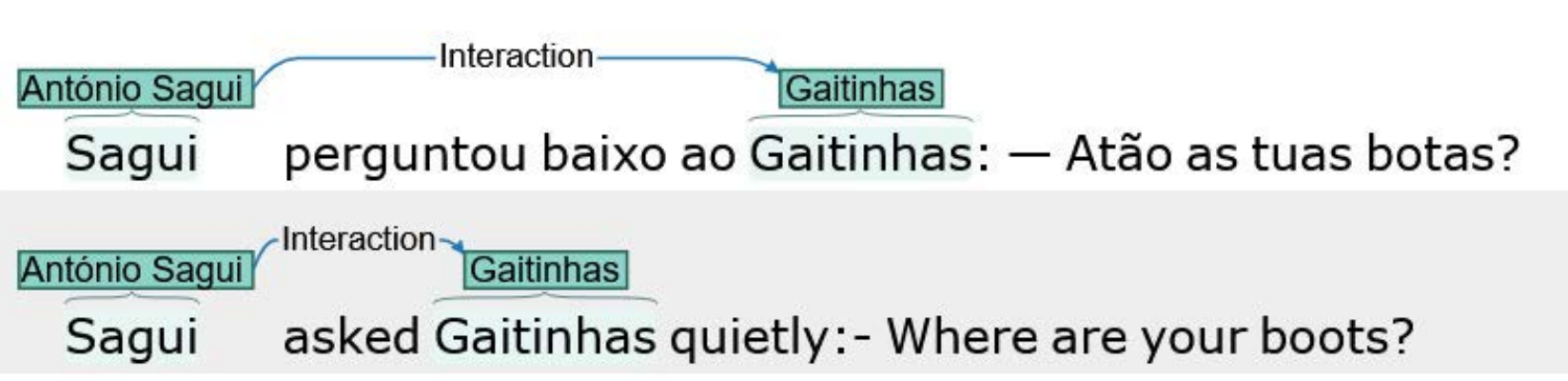}
    \caption{Example of a tagged interaction between two characters.}
    \label{fig:example_characters}
\end{figure}

To benchmark the results of Taggus pipeline, a randomly chosen chapter from each novel was manually tagged. The tagging process consisted of identifying and flagging every instance of a character occurrence in the text. This process considers the possible name variations, thus creating a list of each character and co-references in the tagged chapter. Each time an interaction occurs between two or more characters present in the chapter, the interaction and the involved characters are also tagged, as seen in Figure~\ref{fig:example_characters}, \textit{Sagul} - a reference to the last name of the character \textit{António Sagul} - interacts with \textit{Gatinhas}.

Due to variations in the size of each chapter from book to book, only chapters with a minimum of 1400 tokens per chapter were considered for tagging. From the 11 manually tagged chapters, the smallest one contains 1481 tokens and the largest 5580, in a total of 34904 manually tagged tokens and an average of 3173 tokens per chapter. The size of this new manually annotated corpus is comparable to most work done in Character Network Extraction with limited resources and time constraints preventing larger samples \citep{elson2010social, colladanay2015clustering, jayannavar2015validating}.

\section{Taggus}
This Section describes the proposed pipeline, named Taggus, to create Character Networks from literary novels. 

The Taggus pipeline consists of the following three tasks to perform automatic character network extraction for literary novels \citep{labatut2019extraction}, as established in our Literature Review (Section~\ref{sec:literature_review}):
\begin{enumerate}
\item[A.] Extract Character names and perform Co-Reference Resolution;
\item[B.] Detect interactions between characters;
\item[C.] Create a graphical representation illustrating the interactions between characters extracted from Tasks A and B.
\end{enumerate}

The remainder of this section describes the approaches implemented for each task in the Taggus pipeline.

\subsection{Character names extraction}

As described previously in the Related Work Section~\ref{sec:related_work_network_extraction}, NER tools in languages other than English tend to underperform in Character Names Extraction (CNE) as they return many false positives. Therefore, Taggus CNE follows a Part-of-Speech (POS) tagging approach based on works developed in other languages \citep{trovati2014towards}. POS tagging enables greater supervision and customization, considering the rules specific to each language \citep{krug2015rule}. For instance, titles usually appear before the name (e.g., Sir), but in some cases, they can follow it (e.g., Junior or Esq.).

In particular, Taggus CNE, based in \citep{trovati2014towards,colladanay2015clustering}, combines two different off-the-shelf tools, (i) spaCy's pt-core-news-lg model \citep{pt_core_news_lg,rademaker2017universal}, a large Portuguese language model providing word vector representations for several NLP tasks, like POS tagging and NER; and (ii) LX-Tagger \citep{branco2004pos}, a part-of-speech (POS) tagger for Portuguese. These two tools are used to mitigate the out-of-domain use of these tools for the Portuguese language.

\input{Table2}

Using LX-Tagger, Taggus extracts character names into a list of character names by identifying patterns composed of combinations of specific tagged tokens like \textit{Sr. Domingos} tagged respectively with \textsc{Title} (\textit{Sr.} translates to Sir in Portuguese) and \textsc{Part of Proper Name} (Domingos is a first name in Portuguese). Table~\ref{table:Table2} lists the tag patterns and examples. These tag patterns are based on the work done by Trovati \& Brady \citep{trovati2014towards} and Adanay \& Sporleder \citep{colladanay2015clustering} and are adapted to the rules of Portuguese names.

\begin{figure}[!t]
    \centering
    \includegraphics[width=0.75\columnwidth]{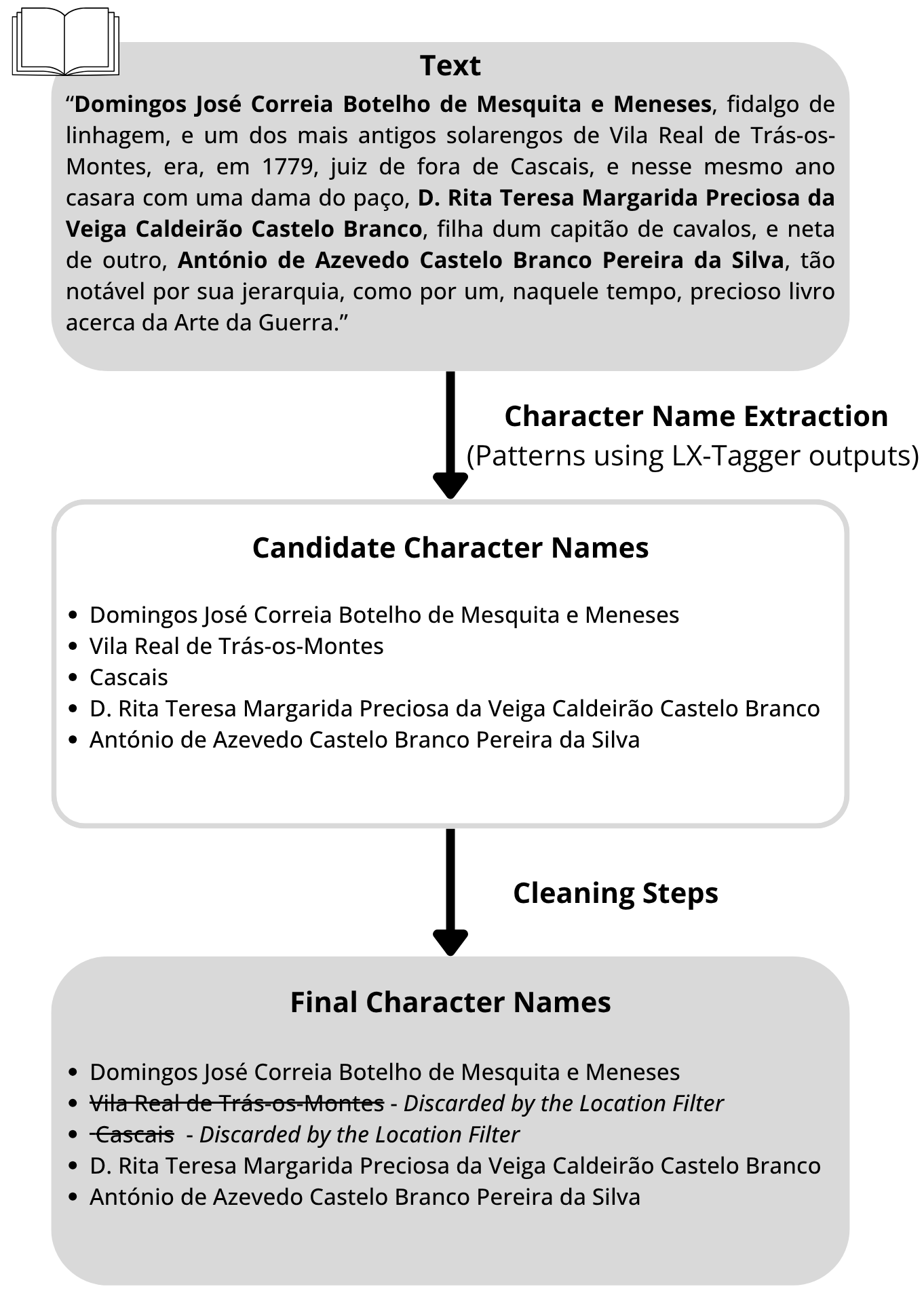}
    \caption{Example of output achieved by the CNE using patterns based on the LX-Tagger POS tags, followed by the steps that conduct the cleaning process.}
    \label{fig3}
\end{figure}

The resulting list of candidate character names contains a series of tokens/words that are not exclusively character names (false positives), like Cascais (a city) or Escorpião de Jade (animal). 
Therefore, Taggus applies the following six steps to clean the candidates to achieve a list that contains exclusively character names, see Figure \ref{fig3} for examples of performing CNE and using the following steps as cleaning filters:

\begin{itemize}

\item \textbf{Title Finder.} The first technique finds any sequence of tokens that contains a title like \textit{Senhora} (Lady) since we have a guarantee this corresponds to a person's name. 

\item \textbf{Presence Detection.} A similar approach to the Title Finder step is performed on verbs or other words that call for the presence of a person indicator like: \textit{gritou, suspirou, perguntou} (shouted, sighed, asked) \citep{freitas2017verbos}. The intuition is that if one of these presence indicators precedes an identified sequence, it is most likely a character \citep{agarwal2012social,trovati2014towards,he2013identification}.

\item \textbf{Spacy Re-tagger.} The remaining sequences on the list are POS re-tagged by spaCy to eliminate incorrect token sequences that LX-Tagger may have flagged as proper names.

\item \textbf{Locations Filter.} We then cross the remaining sequences with a geographic database  \citep{WorldCitiesDatabase} to remove sequences that consist only of cities or locations.

\item \textbf{Incorrect Tokens Removal.} A smaller processing step consists of removing tokens that were incorrectly identified as part of a name by finding lower-case versions of said tokens in the text. For example, if the sentence begins with a common noun such as "Book," the POS taggers in the previous steps recognize it as a proper noun due to its capitalization. However, by implementing this step and identifying the same word in lowercase, the system recognizes that it cannot be a proper noun and removes it from the list. 

\item \textbf{First Names Filter.} Lastly, the remaining working list is checked with a database of Portuguese names \citep{centraldedados_nomes_proprios}. This step filters out full names whose first token/word is not a valid name in the database. For example, for the candidate name sequence \textit{Domingos José Correia}, this filter would keep this sequence as \textit{Domingos} is in the database.
\end{itemize}

The output of this phase, after processing the six steps above, results in a list of character names, including their different name variations used to refer to the same character, e.g, \textit{Sr. Domingos} and \textit{Domingos José Correia}. The next Section describes the co-reference resolution steps that group and normalize character names into a unique representation.

\subsection{Co-reference resolution}
\begin{figure}[!t]
    \centering
    \includegraphics[width=\textwidth]{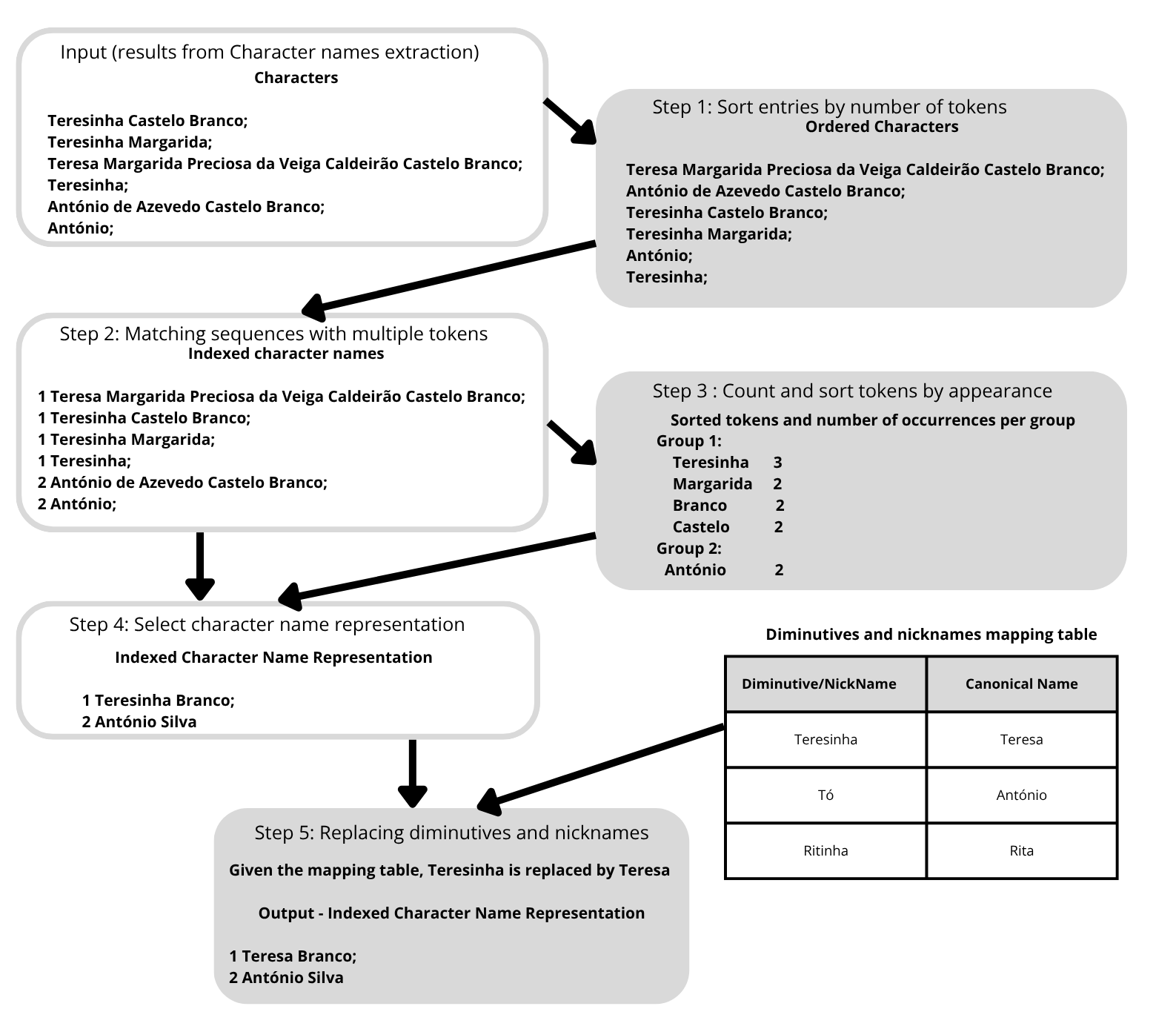}
    \caption{Example of output for each step of the Co-reference resolution process.}
    \label{fig4}
\end{figure}

With the resulting list of clean character names, the pipeline performs co-reference resolution. Our approach consists of a set of six steps. These were achieved by adapting and combining existing methodologies \citep{krug2015rule} \citep{elsner2012kernels} \citep{vala2015mr} \citep{colladanay2015clustering} to Portuguese names. Examples of the output achieved by each step can be seen in Figure \ref{fig4}, and they are as follows:

\begin{enumerate}
    \item Taggus starts by creating a new sequence whose entries are descending by the highest number of tokens.  See \textit{Step 1: Sort Entries by number of tokens}, Figure \ref{fig4};
    \item Taggus starts an iterative process through the new sequence where it picks the first entry and searches each subsequent entry if their tokens match with the tokens of the picked entry. If all subsequent tokens match, it is added to the entry index; if not, a new index is created. This process is repeated until an index is assigned to every entry. See \textit{Step 2: Matching sequences with multiple tokens}, Figure \ref{fig4};
    
    \item Within each group, i.e., entries that share the same index, Taggus counts the occurrences of each token and creates a sorted list.
    See \textit{Step 3: Count and sort tokens by appearance}, Figure \ref{fig4};
    
    \item For each group, a character name representation is selected. It consists of the first and last name from the longest full name that begins with the most frequent group token identified in the previous step. See \textit{Step 4: Select character name representation}.

    \item The character names are then checked for matching in a manually constituted list of diminutives and nicknames, and turned into their canonical form. If a match is found, the groups that contain the same tokens are joined together. See \textit{Step 5: Replacing diminutives and nicknames}, Figure \ref{fig4};
    \item Lastly, the system asks us, the user, if the book is written in the first person, if so, we are asked to indicate the group corresponding to the narrating character \citep{gessey2020narrative}. Therefore, mentions of "I" and "me" are then assigned to the group representation of the narrating character.
\end{enumerate}

This step ends with a list of groups representing one character and all the possible name variations found in the text.
To avoid False Positives, an additional processing step discards every character that occurs less than three times throughout the novel, this helps to clean any previous mistakes made by the system and entries that consist of mentions instead of characters \citep{elsner2012kernels}, e.g. “Luís de Camões”.

Also included in this task, but less relevant to our final results, is a system that extracts information about a character’s gender, which was also set up. This information can later be used to further analyze the novel and its characters, where gender might be a relevant factor \citep{elsner2012kernels}. We achieved this by cross-referencing names with the database of Portuguese names. Those not found in the database are selected through a voting system, where the preceding tokens in the sequence are retrieved and used to determine gender based on the most prominent indicator. This is done by collecting gendered pronouns such as "o" / "a", or possessive markers like "do" / "da". The occurrences for male and female indicators are counted, and the literary character is assigned the corresponding gender.

\subsection{Interactions detection}
Following the Character Name Extraction and Co-reference resolution, Taggus performs the task of detecting interactions between characters. As referenced in the Standard Pipeline description in Section~\ref{sec:related_work_system}, co-occurrence is, with all its flaws, still the current best solution for interaction detection. Capturing both verbal interactions and interactions that occur in the body of the text is the best method when treating novels \citep{colladanay2015clustering}.

Taggus pipeline thus considers that an interaction exists when two characters from the list are mentioned in the same text window. This window has been set to 3 sentences \citep{silva2023analyzing}. This frame was determined to be the most capable of capturing the majority of interactions without returning a large number of False Positives. 

\begin{figure}[!t]
    \centering
    \includegraphics[width=0.6\textwidth]{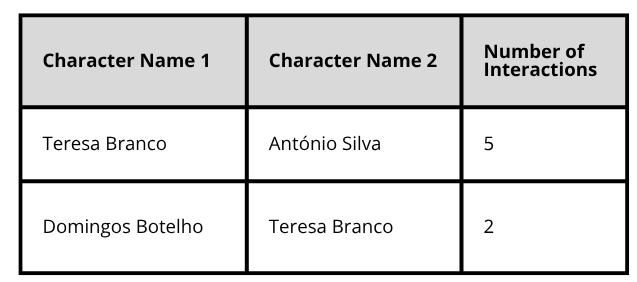}
    \caption{Example of an interaction table.}
    \label{fig:interactions_table}
\end{figure}

This step returns a table with three columns whose first two present the names of characters interacting with each other, and the last column reports the number of interactions between them throughout the book, see Figure \ref{fig:interactions_table}.

\subsection{Generation of the network visualization}
\begin{figure}[!t]
    \centering
    \includegraphics[width=0.75\columnwidth]{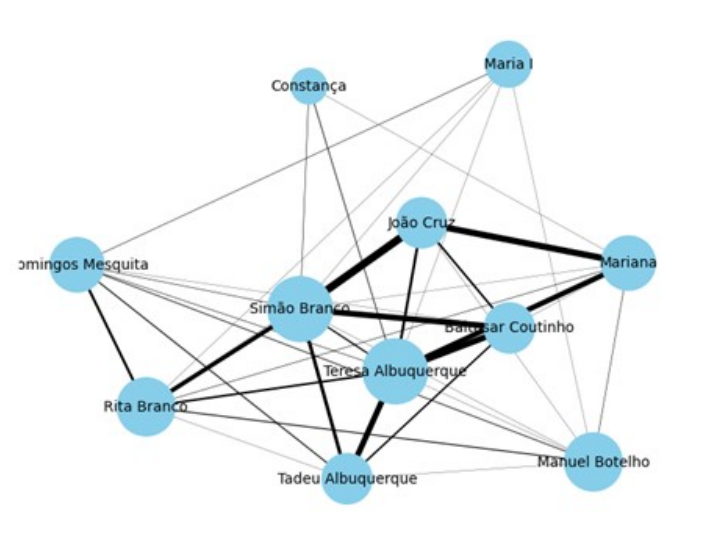}
    \caption{Character Network from “Amor de Perdição”.}
    \label{fig5}
\end{figure}

Using the resulting character names and interactions from the previous two tasks, Taggus computes and plots a static character network, where characters are represented as nodes and interactions as edges. Both nodes and edges are weighted according to their relevance. Taggus outputs graphical representations of the networks; as an example of such a network, Figure \ref{fig5} displays the network from "Amor de Perdição".

\section{Results and Discussion}
This work set out to answer how Character Network Extraction (CNE) methods could be adapted to address the unique linguistic challenges in Portuguese literary works.

To that extent, we developed a CNE pipeline that leverages the available tools for the Portuguese language, combining them with previously tested methods to attempt to reproduce the results achieved by English language SOTA tools utilized in these tasks.

As already established, we can consider the pipeline to consist of three subtasks: (i) identifying character occurrences while performing co-reference resolution, (ii) detecting interactions between said characters, and (iii) the subtask consisting of building a graphical representation of the ensuing results. The first and second subtasks' outputs are the most significant to our final product, and the third subtask is subject to evaluation.

To evaluate Taggus's performance in identifying character occurrences during co-reference resolution and detecting interactions between characters, we benchmark the results on manually tagged chapters and compare Taggus’s pipeline outputs with those of available tools that perform these tasks in Portuguese.

Three metrics are used in our assessment \citep{rocha2014entity}: 
Precision answers off all the instances of characters or interactions found in the text by our pipeline, how many of them are actually correct;
Recall answers of all the existing character names or interactions instances found in text, how many did the pipeline correctly identified;
and F1-Score is the harmonic mean between both Precision and Recall.

The remainder of this Section reports and discusses the results individually for each task in the CNE pipeline and provides a discussion of the overall results.

\subsection{Character names extraction and co-reference resolution}

This Section evaluates the results of the character occurrence detection process and compares them to those obtained from the off-the-shelf Named Entity Recognition tool spaCy’s Portuguese language model “\texttt{pt\_core\_news\_lg}.” (not to be confused with the POS tagging of said model used in the Taggus pipeline). These results provide a better understanding of what the outcome would be achieved by a NLP system without all the further steps we implemented.

The results regarding Character name extraction are present in Table 3.

\input{Table3}

As can be seen, Taggus significantly outperforms the off-the-shelf Spacy's NER model for Portuguese both in Precision and F1-Score, with an average of 93.9\% and 94.1\% against 29.4\% and 43.4\%, respectively. Regarding Recall, the difference is less significant; Taggus achieves 94.5\% as opposed to 92.5\% for the NER tool. We can thus conclude that the NER tool can correctly identify instances of characters in novels, but does not do so without returning a large volume of False Positives. The lower Precision, i.e., excess of false positives, obtained by Spacy's NER tool is expected as the literary texts linguistically differ from the texts used to train such NER models, and the NER tools tend to incorrectly identify tokens as Character Names. In Taggus, we prevent this situation by using specific filters that catch most of such tokens.

Overall, Taggus achieved very promising results in this task. With no other systems developed for the Portuguese language to compare to, our scores indicate that the pipeline was very competent in identifying character names and their variations in the manually tagged samples. The lowest performance, seen for “O Cortiço” with 77.4\% Recall, is explained by the uniqueness of the sample chapter itself, where a total of 27 characters were manually identified, with 15 making a single appearance in the text. This suggests that our pipeline may struggle to identify minor characters.

The lowest Precision was recorded for “O Crime do Padre Amaro” with 81.3\% . This sample chapter had the most significant number of tokens, with a Recall of 96.8\%, our system retrieved almost the entirety of the 94 total instances of characters with their corresponding name variations. However, it returned 20 false positives, such as \textit{India} referring to the country and not to a person, and \textit{Palma de o Ministério de as Obras Públicas} the first token \textit{Palma} refers to a person's name, however the remaining tokens refer to the person\textit{Palma} is from the Minister of Public Works. To this extent, we can assess that Taggus, when performing this task in a full novel as intended, is very capable of returning the correct instances of characters solving for co-reference, but is unable to do so without returning a marginal number of misclassified entries.

\subsection{Interactions detection}
Regarding the second task of correctly detecting interactions between characters in a novel, Taggus was compared with ChatGPT 4.0. This choice was made to compare Taggus' performance against a readily available tool that can perform this task and test the current capabilities of Generative Artificial Intelligence to further develop this field. This Large Language Model used one-shot learning, i.e., it was given an example of a character interaction and was asked to replicate this upon being fed each chapter under analysis. Since this task was meant to test capabilities on interaction detection and not co-reference resolution, the evaluation did not penalize the characters’ name variations when comparing the results with the annotations. The results are as seen in Table~\ref{table:results_interaction}.

\input{Table4}

Taggus outperforms the results obtained from its counterpart. ChatGPT struggles with larger chapters, indicating that it could not perform this task in a full novel. Its performance seems to be highly dependent on the writing style, sometimes presenting high Precision but doing so at the cost of a lower Recall, as we can see in “O Crime do Padre Amaro” with 100\% and 25.0\% respectively, or presenting an acceptable Recall, but doing so with low Precision as in “O Cortiço”, with 64.7\% and 27.5\% respectively.

Taggus assessment has shown itself to be more consistent. With an average of 88.2\% Recall and 65.1\% Precision, Taggus seems to be proficient in correctly identifying instances of characters interacting; it does so with a few false positives.

Taking an in-depth look at this false positives issue in “A queda de um Anjo”, where Precision was 55.6\% and Recall 100\%, we observe that only five interactions were manually identified in the chapter, and all were correctly identified. Taggus, however, identified four additional interactions, all of which correspond to characters that were present in this chapter.

The same can be said for “A Escrava Isaura”, where, once again, there were no false negatives out of the twelve manually tagged interactions. Taggus, again, mistakenly recorded more interactions between the characters present in the text and one instance of a character who was not present. This can be explained by the mention of “S. António”, a religious sanctity that our system incorrectly identified as the character “António”. The rest of the false positives were interactions between characters in the chapter that were not tagged as such manually.

\subsection{General Discussion}
Overall, the proposed pipeline, Taggus, achieved positive results in both the assessed tasks and is an improvement from the currently available tools that can perform these tasks. 
The mistakes we observe in the pipeline can be attributed to the ambiguity of what constitutes an interaction. For instance, when two characters exchange dialogue, our pipeline might identify more interactions than those manually assessed. Seeing that Taggus final goal is to represent graphically a network of characters and how connected they are throughout a novel, this is less significant.

The same can be observed for the lower scores in chapters with few interactions. As in the previously mentioned “A queda de um Anjo” chapter, having a low total number of manually tagged interactions makes one mistake by the pipeline very costly on the Precision score. However, when we think of the end goal of our pipeline, one incorrectly tagged interaction is not as relevant in the end product \citep{colladanay2015clustering}.

Other identified problems, such as religious evocations incorrectly identified as characters, are unfortunately a necessary trade-off. In some books, a relevant character might be identified in the same way as saints, such as “S. Joaneira” in “Crime do Padre Amaro". This serves as a reminder of how complex the task ultimately is.

\section{Conclusions}
Here, we introduce Taggus, a pipeline that automates the Extraction of Characters' Social Networks from Portuguese Literary sources. Taggus leverages several standard techniques used in other languages and manages to improve on results obtained using off-the-shelf and state-of-the-art tools.

By reproducing methodologies tested in other languages and adapting them to the Portuguese language rules, we overcame the difficulty that off-the-shelf NLP tools present while trying to perform this task in Portuguese novels. To our knowledge, this is the first automated Character Network Extraction system for the Portuguese language that can be applied, with no requirement for external annotations, to a vast corpus of novels.

Overall, our pipeline has achieved promising results, with a 94.1\% average F1-Score for identifying character occurrences considering co-reference and a 75.9\% F1-Score for detecting interactions between said characters. Anaphoric resolution for co-reference remains challenging, even for English models with access to more effective tools. Some testing was performed in our pipeline to address this, but no satisfactory results have been achieved. Another aspect that can be improved is co-occurrence as an interaction detection method. Its imprecise nature is known within the field, but it is still accepted as the best available method. Future exploration in this matter is needed.

It is important to note that the lack of an available manually annotated corpus limited our ability to evaluate our system’s performance. Although blindly choosing chapters from a corpus of books is standard practice, our system has yet to be tested on a full set of novels, limiting our conclusions. It is also necessary to note that the availability of freely accessible books limited our corpus. This is a common problem and is why most works developed in Character Network Extraction are conducted in 19th-century literature; as such, Taggus has yet to be tested with more modern and contemporary books, which can have different aspects to consider when solving for co-reference resolution. However, the novel “Esteiros” published in 1941, which Taggus performed effectively with $73.7\%$ of F1-Score, has all the characteristics of modern books regarding character naming. Most characters are referred to by their nicknames rather than their full names, suggesting that Taggus would likely perform similarly in contemporary literature.

For future work, several improvements can be made to the project. One key area is enhancing co-reference resolution, particularly in handling pronouns and anaphoric mentions that may have been overlooked in Taggus pipeline. This also includes improving the linking of references to their correct named entities. Moreover, the pipeline could be tested on a larger and more diverse corpus of Portuguese literature, incorporating more contemporary works. Expanding the dataset to include a larger manually annotated corpus would also allow for more robust benchmarking. Regarding benchmarking, it is crucial to track emerging trends and fine-tune new NLP models for Portuguese language tasks. Comparing Taggus against these new advancements would help assess its performance and identify areas for further improvement. Finally, incorporating interaction directionality would allow for creating a directed graph, further refining the representation of relationships within the network.

To this extent, our pipeline, Taggus, is publicly available \citep{Taggus} not only to allow for studies in Character Networks Analysis but also to encourage further exploration of these issues and progress in work done for the Portuguese language.

\backmatter
%

\bmhead{Acknowledgements}
FLP acknowledges the financial support provided by FCT Portugal under the project UIDB/04152 -- Centro de Investigação em Gestão de Informação (MagIC). JLMP acknowledges the financial support provided by FCT Portugal under the project UIDB/00319 -- Centro de Investigação ALGORITMI.

\section*{Declarations}
\bmhead{Competing interests}
The authors declare no competing interests

\bmhead{Data availability}
Data is available upon reasonable request to the authors.

\bmhead{Code availability}
The code necessary for reproducing the results of the manuscript, including the pipeline implementation, is publicly available on GitHub repository \citep{Taggus}.

\bmhead{Author contribution}
Tiago G. Canário and Catarina Duarte: Methodology, Programming, Formal analysis, Data Curation, Writing - Original Draft, Visualization
Flávio L. Pinheiro and João L. M. Pereira: Conceptualization, Writing - Review \& Editing, and Supervision.


\bibliography{references}

\end{document}

%% file: Table1.tex
\begin{table}[tbp]
\centering
\caption{List of Novels used in this manuscript to benchmark the Taggus pipeline. The table shows the names of the authors and of the titles of the novels.}
\label{tab:list_novels}
\begin{tabular}{llcc}
\hline
\rowcolor[HTML]{EFEFEF} 
\textbf{Author Name}   & \textbf{Title}                     & \textbf{Year} & \textbf{Nationality} \\ \hline
Aluíso Azevedo         & O Cortiço                          & 1890          & BR                   \\
Bernardo Guimarães     & A Escrava Isaura                   & 1875          & BR                   \\
Camilo Castelo Branco  & Amor de Perdição                   & 1862          & PT                   \\
Camilo Castelo Branco  & A Queda dum Anjo                   & 1866          & PT                   \\
Eça de Queiroz         & O crime do padre amaro             & 1875          & PT                   \\
Júlia Lopes de Almeida & A Viúva Simões                     & 1897          & BR                   \\
Júlio Dinis            & As Pupilas do Senhor Reitor        & 1863          & PT                   \\
Lima Barreto           & O Triste Fim de Policarpo Quaresma & 1915          & BR                   \\
Machado de Assis       & Dom Casmurro                       & 1899          & BR                   \\
Mário de Sá-Carneiro   & A Confissão de Lúcio               & 1914          & PT                   \\
Soeiro Pereira Gomes   & Esteiros                           & 1941          & PT                   \\ \hline
\end{tabular}
\end{table}

%% file: Table2.tex
\begin{table}[tbp]
\centering
\caption{Tag patterns for character names extraction using LX-tagger POS tags. The matching example refers to different name representations for the Domingos character that match the given tag pattern. \textit{Sr.} means Sir in Portuguese, \textit{de} is a preposition, the remaining words/tokens present are parts of Proper names.} \label{table:Table2}
\renewcommand{\arraystretch}{1.8}
\begin{tabular}{p{0.5\textwidth} p{0.45\textwidth}}
\hline
\rowcolor[HTML]{EFEFEF} 
\textbf{Tag Pattern}                                                   & \textbf{Matching Example}                     \\ \hline
Proper Name                                                            & Domingos                                      \\
Title + Proper Name                                                    & Sr. Domingos                                  \\
Proper Name + Proper Name                                              & Domingos José Correia Botelho                 \\
Proper Names + Prepositions/Definite articles if + Proper Name         & Domingos José Correia Botelho de Mesquita     \\
Title + Proper Names + Prepositions/Definite articles if + Proper Name & Sr. Domingos José Correia Botelho de Mesquita \\ \hline
\end{tabular}
\end{table}

%% file: Table3.tex
\begin{table*}[!t]
\centering
\caption{Pipeline’s metrics in Character Name Extraction compared to NER. The highest achieved score for each metric is highlighted in bold.
}
\footnotesize
\begin{tabular}{lcccccc}
\hline
\rowcolor[HTML]{EFEFEF} 
\cellcolor[HTML]{EFEFEF} &
  \multicolumn{3}{c}{\cellcolor[HTML]{EFEFEF}\textbf{Taggus}} &
  \multicolumn{3}{c}{\cellcolor[HTML]{EFEFEF}\textbf{Spacy NER}} \\
\rowcolor[HTML]{EFEFEF} 
\multirow{-2}{*}{\cellcolor[HTML]{EFEFEF}\textbf{List of Novels}} &
  Precision &
  Recall &
  F1-Score &
  Precision &
  Recall &
  F1-Score \\ \hline
A Confissão de Lucio             & \textbf{96.6} & \textbf{93.4} & \textbf{95.0} & 50.0 & 84.6 & 62.9 \\
A queda de um Anjo               & \textbf{92.9} & \textbf{96.3} & \textbf{94.5} & 11.1 & \textbf{96.3} & 19.9 \\
A Viuva Simões                   & \textbf{93.9} & \textbf{100}  & \textbf{96.9} & 43.7 & \textbf{100}  & 60.8 \\
Amor de Perdição                 & \textbf{98.3} & \textbf{100}  & \textbf{99.2} & 21.1 & \textbf{100}  & 34.9 \\
As Pupilas do Senhor Reitor      & \textbf{93.7} & \textbf{93.7} & \textbf{93.7} & 27.1 & \textbf{93.7} & 42.0 \\
Dom Casmurro                     & \textbf{98.4} & \textbf{98.4} & \textbf{98.4} & 31.2 & 77.4 & 44.4 \\
Escrava Isaura                   & \textbf{97.8} & 91.7 & \textbf{94.6} & 22.6 & \textbf{93.8} & 36.4 \\
Esteiros                         & \textbf{98.8} & 97.7 & \textbf{98.2} & 32.0 & \textbf{100}  & 48.5 \\
O Cortiço                        & \textbf{90.6} & 77.4 & \textbf{83.5} & 36.8 & \textbf{90.3} & 52.3 \\
O Crime do Padre Amaro           & \textbf{81.3} & \textbf{96.8} & \textbf{88.3} & 15.1 & 87.2 & 25.7 \\
Triste Fim de Policarpo Quaresma & \textbf{90.8} & \textbf{94.7} & \textbf{92.7} & 33.2 & \textbf{94.7} & 49.2 \\ \hline
Average                          & \textbf{93.9} & \textbf{94.5} & \textbf{94.1} & 29.4 & 92.5 & 43.4 \\ \hline
\end{tabular}
\end{table*}

%% file: Table4.tex
\begin{table*}[!t]
\centering
\caption{Pipeline’s metrics in detected interactions compared to ChatGPT. The highest achieved score for each metric is highlighted in bold} \label{table:results_interaction}
\footnotesize
\begin{tabular}{lcccccc}
\hline
\rowcolor[HTML]{EFEFEF} 
\cellcolor[HTML]{EFEFEF} &
  \multicolumn{3}{c}{\cellcolor[HTML]{EFEFEF}\textbf{Taggus}} &
  \multicolumn{3}{c}{\cellcolor[HTML]{EFEFEF}\textbf{ChatGPT 4.0}} \\
\rowcolor[HTML]{EFEFEF} 
\multirow{-2}{*}{\cellcolor[HTML]{EFEFEF}\textbf{List of Novels}} &
  \multicolumn{1}{l}{\cellcolor[HTML]{EFEFEF}Precision} &
  \multicolumn{1}{l}{\cellcolor[HTML]{EFEFEF}Recall} &
  \multicolumn{1}{l}{\cellcolor[HTML]{EFEFEF}F1-Score} &
  \multicolumn{1}{l}{\cellcolor[HTML]{EFEFEF}Precision} &
  \multicolumn{1}{l}{\cellcolor[HTML]{EFEFEF}Recall} &
  \multicolumn{1}{l}{\cellcolor[HTML]{EFEFEF}F1-Score} \\ \hline
A Confissão de Lucio             & \textbf{80.0} & \textbf{88.9} & \textbf{84.2} & 11.4 & 55.6 & 18.9 \\
A queda de um Anjo               & 55.6 & \textbf{100}  & \textbf{94.5} & \textbf{100}  & 40.0 & 57.1 \\
A Viuva Simões                   & \textbf{66.7} & \textbf{100}  & \textbf{80.0} & \textbf{66.7} & 75.0 & 70.6 \\
Amor de Perdição                 & \textbf{81.3} & \textbf{81.3} & \textbf{81.3} & 41.7 & 62.5 & 50.0 \\
As Pupilas do Senhor Reitor      & 75.9 & \textbf{91.7} & \textbf{83.0} & \textbf{83.3} & 62.5 & 71.4 \\
Dom Casmurro                     & \textbf{56.0} & \textbf{87.5} & \textbf{68.3} & 40.6 & 81.3 & 54.2 \\
Escrava Isaura                   & 57.1 & \textbf{100}  & 72.7 & \textbf{90.0} & 75.0 & \textbf{81.8} \\
Esteiros                         & \textbf{73.7} & \textbf{73.7} & \textbf{73.7} & 46.4 & 68.4 & 55.3 \\
O Cortiço                        & \textbf{52.4} & \textbf{64.7} & \textbf{57.9} & 27.5 & \textbf{64.7} & 38.6 \\
O Crime do Padre Amaro           & 48.9 & \textbf{95.8} & \textbf{64.8} & \textbf{100}  & 25.0 & 40.0 \\
Triste Fim de Policarpo Quaresma & \textbf{69.2} & \textbf{87.1} & \textbf{77.1} & 68.4 & 41.9 & 52.0 \\ \hline
Average                          & \textbf{65.1} & \textbf{88.2} & \textbf{76.1} & 61.5 & 59.3 & 53.6 \\ \hline
\end{tabular}
\end{table*}

%% file: main.bbl
\begin{thebibliography}{62}
\bibitem[{Bossaert and Meidert(2013)Bossaert, G. and Meidert,
  N.}]{bossaert2013harry}
Bossaert G, Meidert N.
\newblock "We are only as strong as we are united, as weak as we are divided":
  A dynamic analysis of the peer support networks in the Harry Potter books.
\newblock Open Journal of Applied Sciences. 2013;3(2):174--185.
\newblock \doi{10.4236/ojapps.2013.32024}.
\end{thebibliography}
